\documentclass[11pt,a4paper]{article}

\usepackage[hyperref]{naaclhlt2018}
\usepackage{times}
\usepackage{latexsym}
\usepackage{url}
\usepackage{amsmath}
\usepackage{amssymb}
\usepackage{amsfonts}
\usepackage{graphicx}
\usepackage[labelfont=bf,skip=5pt]{caption}
\usepackage{subcaption}
\usepackage{array}
\usepackage{tabu}
\usepackage{makecell}
\usepackage{subcaption}
\usepackage{paralist}
\usepackage{cases}
\usepackage{diagbox}
\usepackage{enumitem}
\usepackage{soul}
\usepackage{multirow}
\usepackage{verbatim}
\usepackage{tabulary}
\usepackage{booktabs}
\usepackage[mathscr]{euscript}
\usepackage{mathtools}

\usepackage{color}
\usepackage{bm}
\definecolor{orange}{rgb}{1,0.5,0}
\definecolor{mdgreen}{rgb}{0.05,0.6,0.05}
\definecolor{mdblue}{rgb}{0,0,0.7}
\definecolor{dkblue}{rgb}{0,0,0.5}
\definecolor{dkgray}{rgb}{0.3,0.3,0.3}
\definecolor{slate}{rgb}{0.25,0.25,0.4}
\definecolor{gray}{rgb}{0.5,0.5,0.5}
\definecolor{ltgray}{rgb}{0.7,0.7,0.7}
\definecolor{purple}{rgb}{0.7,0,1.0}
\definecolor{lavender}{rgb}{0.65,0.55,1.0}

\newcommand{\ensuretext}[1]{#1}
\newcommand{\marker}[2]{\ensuremath{^{\textsc{#1}}_{\textsc{#2}}}}

\newcommand{\arkcomment}[3]{\ensuretext{\textcolor{#3}{[#1 #2]}}}
\renewcommand{\arkcomment}[3]{}  

\newcommand{\nascomment}[1]{\arkcomment{\marker{NA}{S}}{#1}{blue}}
\newcommand{\hao}[1]{\arkcomment{\marker{H}{P}}{#1}{mdgreen}}
\newcommand{\sam}[1]{\arkcomment{\marker{S}{T}}{#1}{orange}}
\newcommand{\swabha}[1]{\arkcomment{\marker{S}{S}}{#1}{purple}}
\newcommand{\com}[1]{}

\newcommand{\term}[1]{\textbf{#1}} 
\newcommand{\tensor}[1]{\mathbf{#1}}
\newcommand{\seq}[1]{\mathbf{#1}}

\newcommand{\edge}[2]{#1 {\rightarrow} #2}

\newcommand{\bilstm}{\tensor{h}}
\newcommand{\fw}[1]{\overrightarrow{\bilstm}_{#1}}
\newcommand{\bw}[1]{\overleftarrow{\bilstm}_{#1}}

\newcommand{\interalia}[1]{\citep[\emph{inter alia}]{#1}}
\newcommand{\argmax}[1]{\underset{#1}{\operatorname{arg}\,\operatorname{max}}\;}

\newcommand{\fnframe}[1]{\textsc{#1}}
\newcommand{\fnrole}[1]{\textsc{#1}}

\newcommand{\dmtop}{\texttt{top}}
\newcommand{\dmlabel}[1]{\texttt{#1}}

\usepackage{xspace}
\newcommand{\sesame}{\texttt{open-SESAME}}
\newcommand{\lu}{LU\xspace}

\DeclarePairedDelimiter\floor{\lfloor}{\rfloor}

\newcolumntype{L}[1]{>{\raggedright\let\newline\\\arraybackslash\hspace{0pt}}m{#1}}
\newcolumntype{C}[1]{>{\centering\let\newline\\\arraybackslash\hspace{0pt}}m{#1}}
\newcolumntype{R}[1]{>{\raggedleft\let\newline\\\arraybackslash\hspace{0pt}}m{#1}}

\setul{1pt}{.4pt}

\DeclareCaptionFont{tiny}{\tiny}
\aclfinalcopy 
\def\aclpaperid{539} 

\title{Learning Joint Semantic Parsers from Disjoint Data}

\author{
	Hao Peng$^\diamondsuit$ \quad
	Sam Thomson$^\clubsuit$ \quad
	Swabha Swayamdipta$^\clubsuit$ \quad
	Noah A. Smith$^\diamondsuit$ \\
	$^\diamondsuit$ Paul G. Allen School of Computer Science \& Engineering,
	University of Washington,
	Seattle, WA, USA \\
	$^\clubsuit$ School of Computer Science,
	Carnegie Mellon University,
	Pittsburgh, PA, USA \\
	{\tt \{hapeng,nasmith\}@cs.washington.edu,
		\{sthomson,swabha\}@cs.cmu.edu}
}
\date{}

\makeatletter
\renewcommand{\paragraph}{%
  \@startsection{paragraph}{4}%
  {\z@}{0.25ex \@plus 0.5ex \@minus .2ex}{-1em}%
  {\normalfont\normalsize\bfseries}%
}
\makeatother

\begin{document}

\maketitle

\begin{abstract}
We present a new approach to learning semantic parsers from multiple datasets, even when the target semantic formalisms are drastically different, and the underlying corpora do not overlap.
We handle such ``disjoint'' data by treating annotations for unobserved formalisms as latent structured variables.
Building on state-of-the-art baselines, we show improvements both in frame-semantic parsing and semantic dependency parsing by modeling them jointly.
Our code is open-source and available at
\url{https://github.com/Noahs-ARK/NeurboParser}.
\end{abstract}

\section{Introduction}
\label{sec:intro}

Semantic parsing aims to automatically predict formal representations of meaning underlying natural language, and has been useful in question answering \cite{Shen:07}, text-to-scene generation~\cite{Coyne:12}, dialog systems \cite{Chen:13} and social-network extraction \cite{Agarwal:14}, among others.
Various formal meaning representations have been developed corresponding to different semantic theories~\cite{fillmore1982frame,palmer2005propbank,flickinger_deepbank_2012,banarescu_abstract_2013}.
The distributed nature of these efforts results in a set of annotated resources that are similar in spirit, but not strictly compatible.
A major axis of structural divergence in semantic formalisms is whether based on \emph{spans}~\cite{baker1998frame,palmer2005propbank} or \emph{dependencies}~\interalia{surdeanu2008conll,oepen2014sdp,banarescu_abstract_2013,copestake_minimal_2005}.
Depending on application requirements, either might be most useful in a given situation.

Learning from a union of these resources seems promising, since more data almost always translates into better performance.
This is indeed the case for two prior techniques---parameter sharing ~\cite{fitzgerald2015semantic,kshirsagar2015frame}, and joint decoding across multiple formalisms using cross-task factors that score combinations of substructures from each \cite{peng2017deep}.
Parameter sharing can be used in a wide range of multitask scenarios, when there is no data overlap or even any similarity between the tasks \cite{collober2008unified,soggard2016deep}.
But techniques involving joint decoding have so far only been shown to work for parallel annotations of dependency-based formalisms, which are structurally very similar to each other~\cite{lluis2013joint,peng2017deep}.
Of particular interest is the approach of Peng et al., where three kinds of semantic graphs are jointly learned on the same input, using parallel annotations. 
However, as new annotation efforts cannot be expected to use the same original texts as earlier efforts, the utility of this approach is limited.

\begin{figure}
	\centering
	\includegraphics[clip,trim=4.cm 16.5cm 4.cm 10.cm, width=\columnwidth]{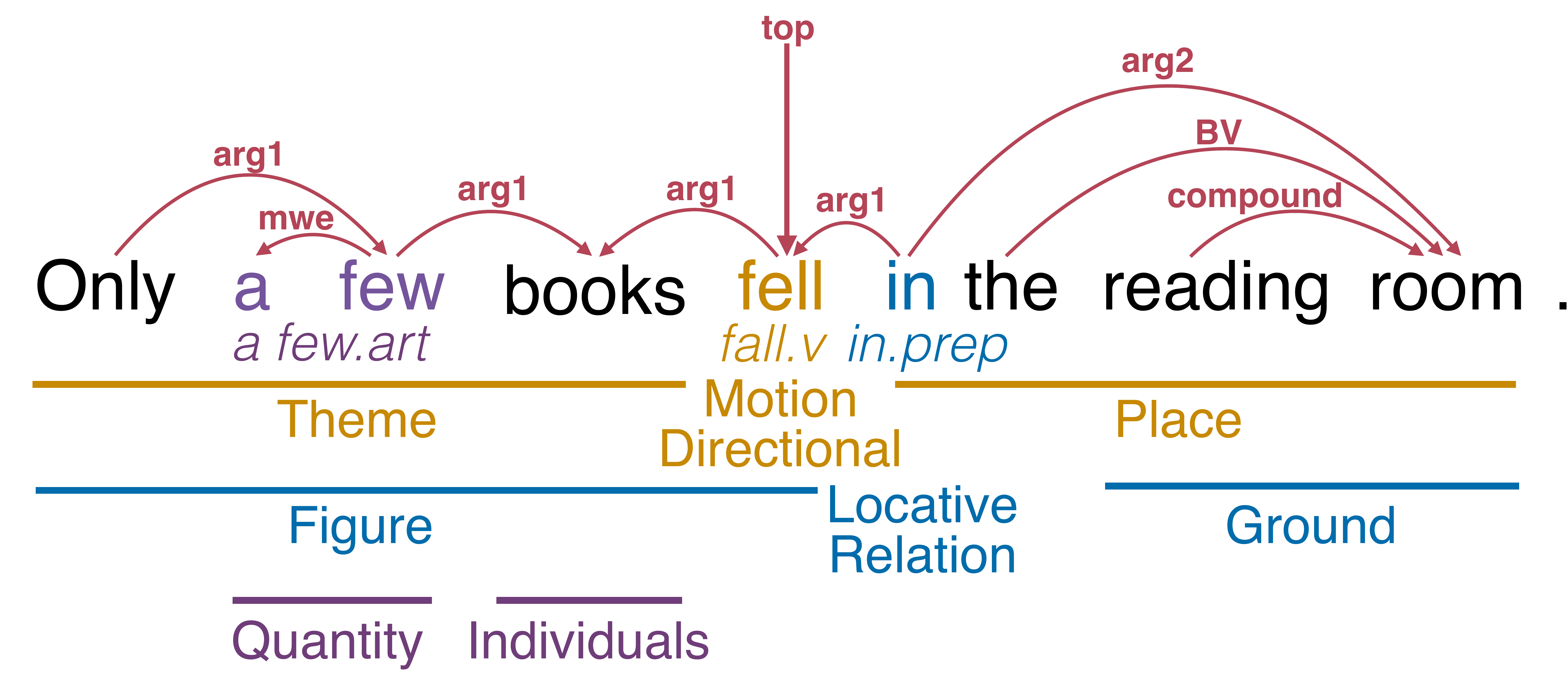}
	\caption{
		An example sentence from the FrameNet 1.5 corpus, shown with an author-annotated DM semantic dependency graph (above) and frame-semantic annotation (below).
		Two more gold frames (and their arguments) have been omitted for space.
	}
	\label{fig:formalisms}
\end{figure}
We propose an extension to Peng et al.'s formulation which addresses this limitation by considering \textbf{disjoint resources}, each containing only a single kind of annotation.
Moreover, we consider \textbf{structurally divergent} formalisms, one dealing with semantic spans and the other with semantic dependencies.
We experiment on frame-semantic parsing~\cite{gildea2002srl,das2010probabilistic}, a span-based semantic role labeling (SRL) task~(\S\ref{subsec:framenet}), and on a dependency-based minimum recursion semantic parsing (DELPH-IN MRS, or DM;~\citealp{flickinger_deepbank_2012}) task~(\S\ref{subsec:sdp}).
See Figure~\ref{fig:formalisms} for an example sentence with gold FrameNet annotations, and author-annotated DM representations. 

Our joint inference formulation handles missing annotations by treating the structures that are not present in a given training example as latent variables (\S\ref{sec:model}).\footnote{Following past work on support vector machines with latent variables \citep{yu2009learning}, we use the term ``latent variable,'' even though the model is not probabilistic.}
Specifically, semantic dependencies are treated as a collection of latent variables when training on FrameNet examples.

Using this latent variable formulation, we present an approach for relating spans and dependencies, by explicitly scoring affinities between pairs of potential spans and dependencies.
Because there are a huge number of such pairs, we limit our consideration to only certain pairs---our design is inspired by the head rules of \citet{surdeanu2008conll}.
Further possible span-dependency pairs are pruned using an $\ell_1$-penalty technique adapted from sparse structure learning (\S\ref{sec:training_inference}).
Neural network architectures are used to score frame-semantic structures, semantic dependencies, as well as cross-task structures~(\S\ref{sec:factors}).

To summarize, our contributions include:
\begin{compactitem}
	\item using a latent variable formulation to extend cross-task scoring techniques to scenarios where datasets do not overlap;
	\item learning cross-task parts across structurally divergent formalisms; and
	\item using an $\ell_1$-penalty technique to prune the space of cross task parts.
\end{compactitem}
Our approach results in a new state-of-the-art in frame-semantic parsing, improving prior work by 0.8\% absolute $F_1$ points~(\S\ref{sec:experiment}), and achieves competitive performance on semantic dependency parsing.
Our code is available at \url{https://github.com/Noahs-ARK/NeurboParser}.

\section{Tasks and Related Work}
\label{sec:task}

We describe the two tasks addressed in this work---frame-semantic parsing~(\S\ref{subsec:framenet}) and semantic dependency parsing~(\S\ref{subsec:sdp})---and discuss how their structures relate to each other~(\S\ref{subsec:head_rule}).

\subsection{Frame-Semantic Parsing}
\label{subsec:framenet}

Frame-semantic parsing is a span-based task, under which certain words or
phrases in a sentence evoke semantic \term{frames}.
A \term{frame} is a group of events, situations, or relationships that all
share the same set of participant and attribute types, called \term{frame elements} or \term{roles}.
Gold supervision for frame-semantic parses comes from the FrameNet lexicon and corpus \cite{baker1998frame}.

Concretely, for a given sentence, $\seq{x}$, a frame-semantic parse $\seq{y}$ consists of:
\begin{compactitem}
  \item a set of \term{targets}, each being a short span (usually a single token\footnote{96.5\% of targets in the training data are single tokens.}) that evokes a frame;
  \item for each target $t$, the \term{frame} $f$ that it evokes; and
  \item for each frame $f$, a set of non-overlapping \term{argument} spans in the sentence, each argument $a = (i, j, r)$ having a start token index $i$, end token
  index $j$ and role label $r$.
\end{compactitem}
The lemma and part-of-speech tag of a target comprise  a \term{lexical unit} (or \term{\lu}). 
The FrameNet ontology provides a mapping from an \lu~$\ell$ to the set of possible frames it could evoke, $\mathcal{F}_\ell$.
Every frame $f \in \mathcal{F}_\ell$ is also associated with a set of roles, $\mathcal{R}_f$ under this ontology. 
For example, in Figure~\ref{fig:formalisms}, the \lu \textit{``fall.v''} evokes the frame \fnframe{Motion\_directional}.
The roles \fnrole{Theme} and \fnrole{Place} (which are specific to \fnframe{Motion\_directional}), are filled by the spans \textit{``Only a few books''} and \textit{``in the reading room''} respectively.
\fnframe{Locative\_relation} has other roles (\fnrole{Profiled\_region}, \fnrole{Accessibility}, \fnrole{Deixis}, etc.) which are not realized in this sentence.

In this work, we assume gold targets and LUs are given, and parse each target independently, following the literature~\interalia{johansson2007lth,fitzgerald2015semantic,yang2017joint,swayamdipta2017frame}.
Moreover, following \citet{yang2017joint}, we perform frame and argument identification jointly.
Most prior work has enforced the constraint that a role may be filled by at most one argument span, but following \citet{swayamdipta2017frame} we do not impose this constraint, requiring only that arguments for the same target do not overlap.

\subsection{Semantic Dependency Parsing}
\label{subsec:sdp}

Broad-coverage \term{semantic dependency parsing} \cite[\term{SDP};][]{oepen2014sdp,oepen2015sdp,Oep:Kuh:Miy:16} represents sentential semantics with labeled bilexical dependencies.
The SDP task mainly focuses on three semantic formalisms, which have been converted to dependency graphs from their original annotations.
In this work we focus on only the DELPH-IN MRS (\term{DM}) formalism.

Each semantic dependency corresponds to a labeled, directed edge between two words.  
A single token is also designated as the \term{\dmtop} of the parse, usually indicating the main predicate in the sentence.
For example in Figure~\ref{fig:formalisms}, the left-most arc has head \textit{``Only''}, dependent \textit{``few''}, and label \dmlabel{arg1}.
In semantic dependencies, the head of an arc is analogous to the target in frame semantics, the destination corresponds to the argument, and the label corresponds to the role.
The same set of labels are available for all arcs, in contrast to the frame-specific roles in FrameNet.

\nascomment{made some changes here to avoid ``head'' and ``modifier'' language}

\subsection{Spans vs. Dependencies}
\label{subsec:head_rule}

Early semantic role labeling was span-based \interalia{gildea2002srl,toutanova2008global}, with spans corresponding to syntactic constituents.
But, as in syntactic parsing, there are sometimes theoretical or practical reasons to prefer dependency graphs.
To this end, \citet{surdeanu2008conll} devised heuristics based on syntactic head rules~\cite{collins2003head} to transform PropBank~\cite{palmer2005propbank} annotations into dependencies.
Hence, for PropBank at least, there is a very direct connection (through syntax) between spans and dependencies.

For many other semantic representations, such a direct relationship might not be present.
Some semantic representations are designed as graphs from the start \citep{hajic2012psd,banarescu_abstract_2013}, and have no gold alignment to spans.
Conversely, some span-based formalisms are not annotated with syntax \citep{baker1998frame,He2015QuestionAnswerDS},%
\footnote{
In FrameNet, phrase types of arguments and their grammatical function in relation to their target have been annotated. 
But in order to apply head rules, the \emph{internal} structure of arguments (or at least their semantic heads) would also require syntactic annotations.}
and so head rules would require using (noisy and potentially expensive) predicted syntax.

Inspired by the head rules of \citet{surdeanu2008conll}, we design cross-task parts, without relying on gold or predicted syntax (which may be either unavailable or error-prone) or on heuristics.

\section{Model}
\label{sec:model}

Given an input sentence $\seq{x}$, and target $t$ with its \lu~$\ell$, 
denote the set of valid frame-semantic parses~(\S\ref{subsec:framenet}) as $\mathcal{Y}(\seq{x}, t,
\ell)$, and valid semantic dependency parses as
$\mathcal{Z}(\seq{x})$.\footnote{For simplicity, we consider only a single
  target here; handling of multiple targets is discussed in \S\ref{sec:experiment}.}
We learn a parameterized function $S$ that scores candidate parses.
Our goal is to \textit{jointly} predict a frame-semantic parse and a semantic dependency graph
by selecting the highest scoring candidates:
\begin{equation}
\label{eq:joint}
\left(\hat{\tensor{y}}, \hat{\tensor{z}}\right) = 
\hspace{-.3cm}
\argmax{(\tensor{y},\tensor{z})\in \mathcal{Y}(\seq{x},t,\ell)\times\mathcal{Z}(\seq{x})}
{\hspace{-.3cm}S(\tensor{y}, \tensor{z}, \tensor{x}, t, \ell)}.
\end{equation}

The overall score $S$ can be decomposed into the sum of frame SRL score $S_{\text{f}}$, semantic dependency score $S_{\text{d}}$, and a cross-task score $S_{\text{c}}$:
\begin{equation}
\begin{split}
S(\seq{y}, \seq{z}, \seq{x}, t, \ell) 
= S_{\text{f}}(\seq{y}, \seq{x}, t, \ell) + S_{\text{d}}(\seq{z}, \seq{x}) \\
+ S_{\text{c}}(\seq{y}, \seq{z}, \seq{x}, t, & \ell).
\end{split}
\end{equation}
$S_{\text{f}}$ and $S_{\text{c}}$ require access to the target and \lu, in addition to $\seq{x}$, but $S_{\text{d}}$ does not.
For clarity, we omit the dependence on
the input sentence, target, and lexical unit, whenever the context is clear.
Below we describe how each of the scores is computed based on the individual \term{parts} that make up the candidate parses.

\paragraph{Frame SRL score.}
The score of a frame-semantic parse consists of
\begin{compactitem}
	\item the score for a predicate part, $s_f(p)$ where each predicate is defined as a combination of a target $t$, the associated \lu, $\ell$, and the frame evoked by the \lu, $f \in \mathcal{F}_{\ell}$;
	\item the score for argument parts, $s_f(a)$, each associated with a token span and semantic role from $\mathcal{R}_{f}$.
\end{compactitem}
Together, this results in a set of frame-semantic parts of size $O(n^2\left|\mathcal{F}_{\ell}\right|\left|\mathcal{R}_f\right|)$.%
\footnote{With pruning (described in \S\ref{subsec:pruning}) we reduce this to a number of parts linear in $n$.
Also, $|\mathcal{F}_{\ell}|$ is usually small (averaging 1.9), as is $\left|\mathcal{R}_f\right|$ (averaging 9.5).
}
The score for a frame semantic structure $\seq{y}$ is the sum of local scores of parts in $\seq{y}$:
\begin{equation}
\label{eq:score_srl}
S_{\text{f}}(\seq{y}) = \sum_{y_i\in \seq{y}} s_{\text{f}}( y_i).
\end{equation}
The computation of $s_\text{f}$ is described in \S\ref{subsec:framenet_score}.

\paragraph{Semantic dependency score.}
Following~\citet{martins2014sdp}, we consider three types of parts in a semantic dependency graph:
semantic heads, unlabeled semantic arcs, and labeled semantic arcs.
Analogous to Equation~\ref{eq:score_srl}, 
the score for a dependency graph $\seq{z}$ is the sum of local scores:
\begin{equation}
\label{eq:score_sdp}
S_{\text{d}}(\seq{z}) = \sum_{z_j\in \seq{z}} s_{\text{d}}(z_j),
\end{equation}
The computation of $s_\text{d}$ is described in \S\ref{subsec:dependency_scoring}.

\begin{figure}
	\centering
	\includegraphics[clip,trim=3.cm 7.5cm 3.cm 9cm, width=\columnwidth]{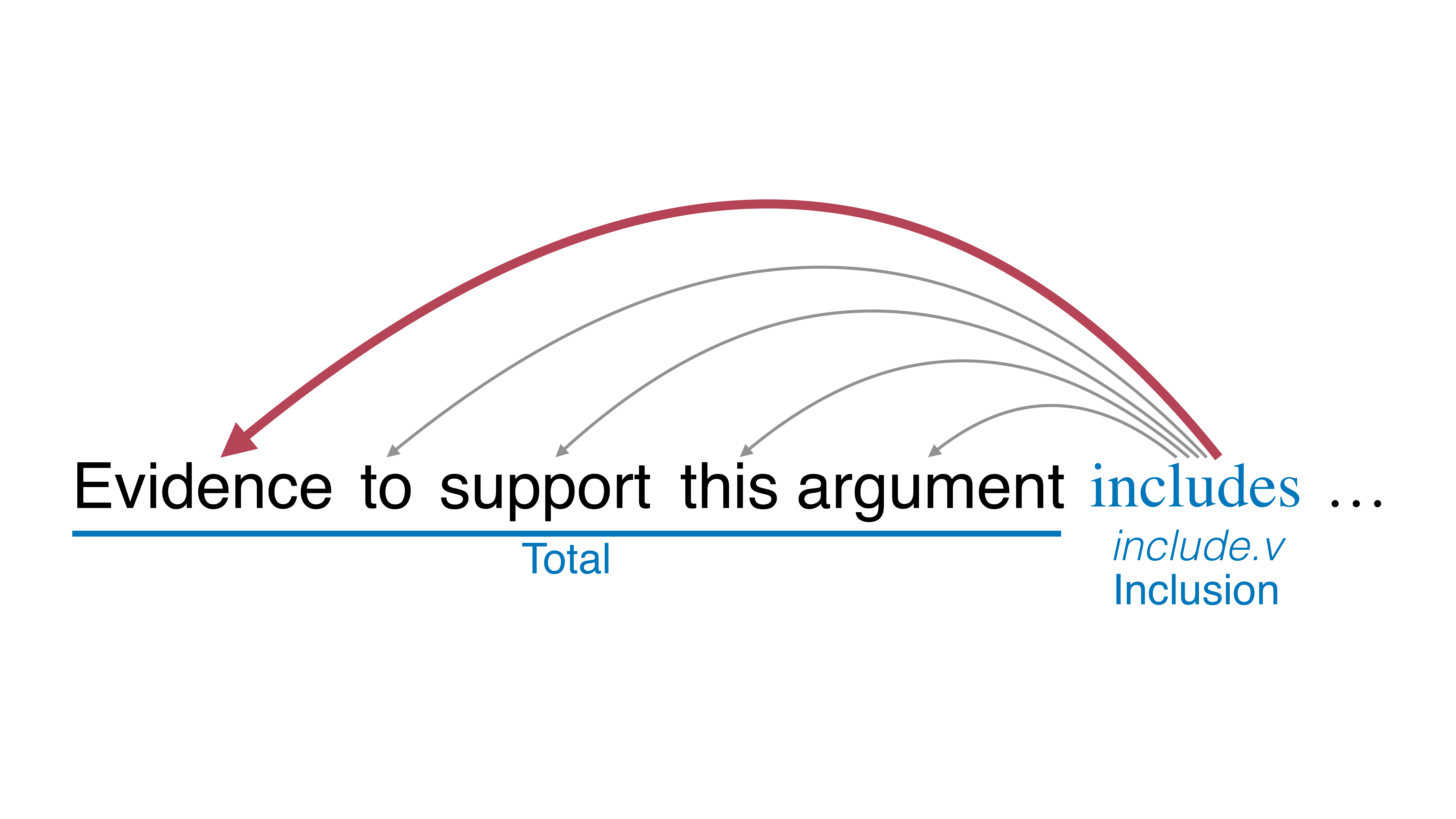}
	\caption{An example of cross-task parts from the FrameNet 1.5 development set. 
		We enumerate all unlabeled semantic dependencies from the first word of the target (\textit{includes}) to any token inside the span.
		The red bolded arc indicates the prediction of our model.}
	\label{fig:ctf}
\end{figure}

\paragraph{Cross task score.}
\label{par:cross-task}
In addition to task-specific parts, we introduce a set $\mathcal{C}$ of cross-task parts.
Each cross-task part relates an argument part from $\seq{y}$ to an \textit{unlabeled} dependency arc from~$\seq{z}$.
Based on the head-rules described in \S\ref{subsec:head_rule}, 
we consider unlabeled arcs from the target to any token inside the span.\footnote{Most targets are single-words (\S\ref{subsec:framenet}). For multi-token targets, we consider only the first token, which is usually content-bearing.}
Intuitively, an argument in FrameNet \emph{would} be converted into a dependency from its target to the semantic head of its span.
Since we do not know the semantic head of the span, we consider all tokens in the span as potential modifiers of the target. 
Figure~\ref{fig:ctf} shows examples of cross-task parts.
The cross-task score is given by
\begin{equation}
\label{eq:cross_task_score}
S_{\text{c}}(\seq{y}, \seq{z})= \sum_{(y_i,z_j)\in(\mathbf{y}\times\mathbf{z})\cap\mathcal{C}}
s_{\text{c}}(y_i, z_j).
\end{equation}
The computation of $s_\text{c}$ is described in \S\ref{subsec:cross_task_scoring}.

In contrast to previous work~\citep{lluis2013joint,peng2017deep},
where there are parallel annotations for all formalisms, 
our input sentences contain only one of the two---either the span-based frame SRL annotations, 
or semantic dependency graphs from DM.
To handle missing annotations, we treat semantic dependencies $\seq{z}$ as latent when decoding frame-semantic structures.\footnote{Semantic dependency parses over a sentence are not constrained to be identical for different frame-semantic targets.}
Because the DM dataset we use does not have target annotations, we do
not use latent variables for frame semantic structures when predicting
semantic dependency graphs.  The parsing problem here reduces to
\begin{equation}
\hat{\tensor{z}} = 
\argmax{\tensor{z}\in \mathcal{Z}}
S_{\text{d}}(\tensor{z}),
\end{equation}
in contrast with Equation~\ref{eq:joint} .

\section{Parameterizations of Scores}
\label{sec:factors}

This section describes the parametrization of the scoring functions from \S\ref{sec:model}.
At a very high level:
we learn contextualized token and span vectors using a bidirectional LSTM \citep[biLSTM;][]{graves2012supervised} and 
multilayer perceptrons (MLPs)~(\S\ref{subsec:reps});
we learn lookup embeddings for LUs, frames, roles, and arc labels;
and to score a part, we combine the relevant representations into a single scalar score
using a (learned) low-rank multilinear mapping.
Scoring frames and arguments is detailed in \S\ref{subsec:framenet_score}, that of dependency structures in \S\ref{subsec:dependency_scoring}, and \S\ref{subsec:cross_task_scoring} shows how to capture interactions between arguments and dependencies.
All parameters are learned jointly, through the optimization of a multitask objective~(\S\ref{sec:training_inference}).

\paragraph{Tensor notation.}
The \term{order} of a tensor is the number of its dimensions---an order-2 tensor is a matrix and an order-1 tensor is a vector.
Let $\otimes$ denote tensor product; the tensor product of two order-2 tensors $\bm{\mathscr{A}}$ 
and $\bm{\mathscr{B}}$
yields an order-$4$ tensor where 
$(\bm{\mathscr{A}} \otimes \bm{\mathscr{B}})_{i,j,k,l}=\bm{\mathscr{A}}_{i,j}\bm{\mathscr{B}}_{k,l}$.
We use $\langle\cdot,\cdot \rangle$ to denote inner products.

\subsection{Token and Span Representations}
\label{subsec:reps}
The representations of tokens and spans are formed using biLSTMs followed by MLPs.

\paragraph{Contextualized token representations.}
Each token in the input sentence $\seq{x}$ is mapped to an embedding vector.
Two LSTMs \citep{hochreiter_long_1997} are run in opposite directions over the input vector sequence.
We use the concatenation of the two hidden representations at each position $i$ as a
contextualized word embedding for each token:
\begin{equation}
\bilstm_i = \bigl[\fw{i}; \bw{i}\bigr].
\end{equation}

\paragraph{Span representations.}
Following \citet{lee2017end2end}, span representations are computed based on 
boundary word representations and discrete length and distance features.
Concretely, given a target $t$ and its associated argument $a = (i, j, r)$ with boundary indices $i$ and $j$, we compute three features $\bm{\phi}_t(a)$ based on the length of $a$, and the distances from $i$ and $j$ to the start of $t$.
We concatenate the token representations at $a$'s boundary
with the discrete features $\bm{\phi}_t(a)$.
We then use a two-layer $\tanh$-MLP to compute the span representation:
\begin{equation}
\mathbf{g}^{\text{span}}(i, j) = \operatorname{MLP}^{\text{span}}{
\bigl(\left[
\bilstm_i;
\bilstm_j; \bm{\phi}_t(a)\right]\bigr)}.
\end{equation}
The target representation $\mathbf{g}^{\text{tgt}}(t)$ is similarly computed using a separate 
$\operatorname{MLP}^{\text{tgt}}$,
with a length feature but no distance features.

\subsection{Frame and Argument Scoring}
\label{subsec:framenet_score}
As defined in \S\ref{sec:model}, the representation for a predicate part incorporates representations of a target span, the associated \lu  and the frame evoked by the \lu. 
The score for a predicate part is given by a multilinear mapping:
\begin{subequations}
\label{eq:pred}
\begin{align}
    \mathbf{g}^{\text{pred}}(f) &=
    \mathbf{g}^{\text{fr}}(f) \otimes \mathbf{g}^{\text{tgt}}(t) \otimes \mathbf{g}^{\text{lu}}(\ell)
  \label{eq:pred_tensor}\\
    s_\text{f}(p)&= \bigl\langle
      \bm{\mathscr{W}},
      \mathbf{g}^{\text{pred}}(f)
    \bigr\rangle,
  \label{eq:pred_score}
\end{align}
\end{subequations}
where $\bm{\mathscr{W}}$ is a 
low-rank order-3 tensor of learned parameters,
and $\mathbf{g}^{\text{fr}}(f)$
and $\mathbf{g}^{\text{lu}}(\ell)$ are learned lookup embeddings for 
the frame and \lu.

A candidate argument consists of a span and its role label, which in turn depends on the frame, target and \lu.
Hence the score for argument part, $a = (i, j, r)$ is given by extending definitions from Equation~\ref{eq:pred}:
\begin{subequations}
  \label{eq:arg}
  \begin{align}
  \mathbf{g}^{\text{arg}}(a) &= 
    \mathbf{g}^{\text{span}}(i, j)\otimes\mathbf{g}^{\text{role}}(r),
    \label{eq:arg_tensor}\\
  s_\text{f}(a) &= \bigl\langle \bm{\mathscr{W}} \otimes \bm{\mathscr{U}}, 
  \mathbf{g}^{\text{pred}}(f) \otimes \mathbf{g}^{\text{arg}}(a)\bigr\rangle,
  \label{eq:arg_score}
  \end{align}
\end{subequations}
where $\bm{\mathscr{U}}$ is a
low-rank order-2 tensor of learned parameters and
$\mathbf{g}^{\text{role}}(r)$ is a learned lookup embedding of the role label. 



\subsection{Dependency Scoring}
\label{subsec:dependency_scoring}
Local scores for dependencies are implemented with two-layer $\tanh$-MLPs,
followed by a final linear layer reducing the represenation to a single scalar score.
\sam{we should explain labeled arc scoring here instead of
  unlabeled. it's easier to make the jump from labeled to unlabeled.}
\nascomment{I agree}
\hao{eq. 11a is reused afterwards. we need to write out $\mathbf{g}^{\text{ua}}$ anyways.
	let's keep unlabeled one. i added some clarifications.}
For example, let $u = \edge{i}{j}$ denote an unlabeled arc (ua).
Its score is:
\begin{subequations}
\label{eqn:s_d}
\begin{align}
  \mathbf{g}^{\text{ua}}(u) &=
    \operatorname{MLP}^{\text{ua}}\bigl(\left[\bilstm_i; \bilstm_j\right]\bigr)
    \label{eq:s_d:g_u} \\
  s_\text{d}(u) &=
    \mathbf{w}^{\text{ua}}\cdot\mathbf{g}^{\text{ua}}(u),
    \label{eq:s_d:s_u}
\end{align}
\end{subequations}
where~$\mathbf{w}^{\text{ua}}$ is a vector of learned weights.
The scores for other types of parts are computed similarly, but with separate~MLPs and weights.

\subsection{Cross-Task Part Scoring}
\label{subsec:cross_task_scoring}

As shown in Figure~\ref{fig:ctf}, each cross-task part $c$ consists of two first-order parts:
a frame argument part $a$, and an unlabeled dependency part, $u$.
The score for a cross-task part incorporates both:
\begin{equation}
\begin{split}
\hspace{-.2cm}
s_\text{c}\left(c\right)
= \bigl\langle \bm{\mathscr{W}} \otimes \bm{\mathscr{U}} \otimes \bm{\mathscr{V}}, \bigr.
&\bigl.\mathbf{g}^{\text{pred}}(f) \otimes \mathbf{g}^{\text{arg}}(a)\bigr.\\
&\bigl.\otimes\ \mathbf{w}^{\text{ua}}\otimes\mathbf{g}^{\text{ua}}(u)\bigr\rangle,
\end{split}
\end{equation}
\sam{$\mathbf{w}^{\text{ua}}$ looks out of place here. are we sure that's right?}
\hao{yes, this is also what we did with arcs in the acl paper}
where $\bm{\mathscr{V}}$ is a
low-rank order-2 tensor of parameters.
Following previous work~\citep{lei2014lowrank,peng2017deep}, we construct the
parameter tensors $\bm{\mathscr{W}}$, $\bm{\mathscr{U}}$, and
$\bm{\mathscr{V}}$ so as to upper-bound their ranks.

\com{
\paragraph{Frame component representations.}
The representations for frames, LUs, and role labels 
are learned lookup embeddings.
\sam{I'm sweeping the affine transformations under the rug,
because they're not actually doing anything.}
}

\section{Training and Inference}
\label{sec:training_inference}

All parameters from the previous sections are trained using a max-margin training objective (\S\ref{subsec:training}).
For inference, we use a linear programming procedure, and a sparsity-promoting penalty term for speeding it up (\S\ref{subsec:inference}).
\subsection{Max-Margin Training}
\label{subsec:training}

Let $\seq{y}^{\ast}$ denote the gold frame-semantic parse,
and let $\delta\left(\seq{y}, \seq{y}^{\ast}\right)$
denote the cost of predicting~$\seq{y}$ with respect to~$\seq{y}^{\ast}$.
We optimize the 
latent structured hinge loss \citep{yu2009learning}, which gives
a subdifferentiable upper-bound on $\delta$:
\begin{equation}
\label{eq:hinge}
\begin{split}
\mathcal{L}\left(\seq{y}^{\ast}\right) =
\hspace{-.3cm}\max_{(\seq{y},\seq{z})\in{\mathcal{Y}\times\mathcal{Z}}}
\hspace{-.1cm}\left\{S\left(\seq{y}, \seq{z}\right) 
+ \delta\left(\seq{y}, \seq{y}^{\ast}\right)\right\}\\
-\max_{\seq{z}\in{\mathcal{Z}}}\left\{S\left(\seq{y}^{\ast}, \seq{z}\right)\right\}.
\end{split}
\end{equation}
Following \citet{martins2014sdp}, 
we use a weighted Hamming distance as the cost function,
where, to encourage recall, we use costs 0.6 for false negative predictions and 0.4 for false positives.
Equation~\ref{eq:hinge} can be evaluated by applying the same max-decoding
algorithm twice---once with cost-augmented inference~\citep{crammer2006online},
and once more keeping $\seq{y}^{\ast}$ fixed.
Training then aims to minimize the average loss over all training instances.
\footnote{We do not use latent frame structures when decoding semantic dependency graphs~(\S\ref{sec:model}).
Hence, the loss reduces to structured hinge~\citep{tsochantaridis2004svm} when training on semantic dependencies.
}

Another potential approach to training a model on disjoint data would be
to marginalize out the latent structures and optimize the conditional
log-likelihood~\citep{naradowsky2012improving}.
Although max-decoding and computing marginals are both NP-hard in general graphical models,
there are more efficient off-the-shelf implementations 
for approximate max-decoding, hence, we adopt a max-margin formulation.


\subsection{Inference}
\label{subsec:inference}
We formulate the maximizations in Equation~\ref{eq:hinge} as 0--1 integer linear
programs and use AD$^3$ to solve them \citep{martins2011dual}.
We only enforce a non-overlapping constraint when decoding FrameNet structures,
so that the argument identification subproblem can be efficiently solved by a dynamic program ~\citep{kong2015segrnn,swayamdipta2017frame}.
When decoding semantic dependency graphs, 
we enforce the determinism constraint~\citep{flanigan2014amr},
where certain labels may appear on at most one arc outgoing from the same token.

\paragraph{Inference speedup by promoting sparsity.}
As discussed in \S\ref{sec:model}, even after pruning, 
the number of within-task parts is linear in the length of the input sentence,
so the number of cross-task parts is quadratic.
This leads to potentially very slow inference.
We address this problem by imposing an $\ell_1$ penalty on the cross-task part scores:
\begin{equation}
\mathcal{L}\bigl(\seq{y}^{\ast}\bigr) + 
\lambda\hspace{-.3cm}\sum_{\left(y_i,z_j\right)\in\mathcal{C}}
\hspace{-.3cm}\bigl\lvert s_c(y_i, z_j)\bigr\rvert,
\end{equation}
where $\lambda$ is a hyperparameter, set to $0.01$ as a practical tradeoff
between efficiency and development set performance.
Whenever the score for a cross-task part is driven to zero, that part's score no
longer needs to be considered during inference.
It is important to note that by promoting sparsity this way,
we do not prune out any candidate solutions.
We are instead encouraging fewer terms in the scoring function, 
which leads to smaller, faster inference problems even though the space of
feasible parses is unchanged.

The above technique is closely related to 
a line of work in estimating the structure of sparse graphical models~\citep{yuan2007model,friedman2008sparse}, where an $\ell_1$ penalty is applied to the inverse covariance matrix in
order to induce a smaller number of conditional dependencies between variables.
To the best of our knowledge, we are the first to apply this technique to the
output of neural scoring functions.
Here, we are interested in learning sparse graphical models only
because they result in faster inference, not because we have any \textit{a priori}
belief about sparsity.
This results in roughly a $14\times$ speedup in our experiments, without any significant
drop in performance. 

\section{Experiments}
\label{sec:experiment}

\begin{table}
	\centering
  \begin{tabulary}{\columnwidth}{@{}lcccc@{}}
    \toprule
 & Train & Exemplars&  Dev. & Test \\ \midrule
FN 1.5 &17,143 & 153,952  & 2,333 & 4,457\\
FN 1.7 & 19,875 &192,460  &  2,308 & 6,722 \\
\midrule
DM id & 33,961 & - & 1,692 &  1,410 \\
DM ood & - & - & - & 1,849  \\
\bottomrule
\end{tabulary}
\caption{\label{tab:data} Number of instances in datasets.}
\vspace{-.5cm}
\end{table}

\paragraph{Datasets.}
Our model is evaluated on two different releases of FrameNet: FN 1.5
and FN 1.7,\footnote{\url{https://FN.icsi.berkeley.edu/fndrupal/}}
using splits from \citet{swayamdipta2017frame}.
Following \citet{swayamdipta2017frame} and \citet{yang2017joint}, 
each target annotation is treated as a separate training instance.
We also include as training data the exemplar sentences, each annotated for a single target, as they have been reported to improve performance~\citep{kshirsagar2015frame,yang2017joint}. 
For semantic dependencies, we use the English DM dataset from the SemEval 2015 Task 18 closed track \citep{oepen2015sdp}.\footnote{\url{http://sdp.delph-in.net/}. 
	The closed track does not have access to any syntactic analyses.
	The impact of syntactic features on SDP performance is extensively studied in \citet{ribeyre2015because}. }
DM contains instances from the WSJ corpus for training and both
in-domain (id) and out-of-domain (ood) test sets, the latter from the
Brown corpus.\footnote{Our FN training data does not overlap with the DM test set.
We remove the 3 training sentences from DM which appear in FN test data.}
Table~\ref{tab:data} summarizes the sizes of the datasets.

\paragraph{Baselines.}
We compare FN performance of our joint learning model (\textsc{Full}) to two baselines:
\begin{compactitem}
	\item[\textbf{\textsc{Basic}:}] A single-task frame SRL model, trained using a structured hinge objective.
	\item[\textbf{\textsc{NoCTP}:}] A joint model without cross-task parts. 
	It demonstrates the effect of sharing parameters in word embeddings and LSTMs (like in \textsc{Full}). 
	It does not use latent semantic dependency structures,
	and aims to minimize the sum of training losses from both tasks.
\end{compactitem} 
We also compare semantic dependency parsing performance
against the single task model by~\citet{peng2017deep}, denoted as NeurboParser (\textsc{Basic}).
To ensure fair comparison with our \textsc{Full} model, 
we made several modifications to their implementation (\S\ref{subsec:impl}).
We observed performance improvements from our reimplementation,
which can be seen in Table~\ref{tab:sdp}.

\paragraph{Pruning strategies.}
\label{subsec:pruning}
For frame SRL, we discard argument spans longer than 20 tokens~\citep{swayamdipta2017frame}.
We further pretrain an unlabeled model and prune spans with posteriors lower than $1/n^2$,  
with $n$ being the input sentence length. 
For semantic dependencies, we generally follow~\citet{martins2014sdp}, replacing their feature-rich pruner with neural networks. 
We observe that $O(n)$ spans/arcs remain after pruning,
with around 96\% FN development recall, and more than 99\% for DM.\footnote{On average, around $0.8n$ argument spans, and $5.7n$ unlabeled dependency arcs remain after pruning.}

\subsection{Empirical Results}
\paragraph{FN parsing results.}
\begin{table}[tb]
	\center
		\begin{tabulary}{\columnwidth}{@{}l   rr  r@{}}
			\toprule
			
			\textbf{Model}
			& \textbf{Prec.}
			& \textbf{Rec.}
			& \textbf{$\bm{F_1}$}\\
			
			\midrule
			
			
			
			Roth
			& 72.2 & 68.0
			& 70.0 \\
			
			T{\"a}ckstr{\"{o}}m
			& 75.4 & 65.8
			& 70.3\\
			
			FitzGerald
			& 74.8 & 65.5
			& 69.9 \\
			FitzGerald ($10\times$)
			& 75.0 & 67.3 
			& 70.9 \\
			
			\sesame
			& 71.0 & 67.8 
			& 69.4 \\
			\sesame~($5\times$)
			& 71.2 & 70.5 
			& 70.9 \\

			Yang and Mitchell (\textsc{Rel})
			& 77.1 & 68.7 
			& 72.7 \\
			
			$^{\dagger\ast}$Yang and Mitchell (\textsc{All})
			& 78.8 & 74.5 
			& 76.6 \\
			
			\midrule[.03em]

			$^\dagger$This work (\textsc{Full})
			& 80.4 & 73.5
			& 76.8 \\
			
			$^\dagger$This work (\textsc{Full}, $2\times$)
			& \textbf{80.4} & \textbf{74.7}
			&  \textbf{77.4} \\
			
			\midrule[.03em]
			$^\dagger$This work (\textsc{Basic})
			& 79.2 & 71.7
			& 75.3\\
			
			$^\dagger$This work (\textsc{NoCTP})
			& 76.9 &  74.8
			& 75.8 \\
			
			\bottomrule
		\end{tabulary}
		\caption{FN 1.5 full structure extraction test performance. 
			$\dagger$ denotes the models jointly predicting frames and arguments, and other systems implement two-stage pipelines and use the algorithm by \citet{hermann2014semantic} to predict frames.
			$K\times$ denotes a product-of-experts ensemble of $K$ models.
			$^\ast$Ensembles a sequential tagging CRF and a relational model. 
			Bold font indicates best performance among all systems.
		}
		\vspace{-.4cm}
		\label{tab:res_15}
\end{table}

Table~\ref{tab:res_15} compares our full frame-semantic parsing results to previous systems.
Among them, \citet{tackstrom2015efficient} and \citet{roth16improving}
implement a two-stage pipeline and use the method from
\citet{hermann2014semantic} to predict frames.
\citet{fitzgerald2015semantic} uses the same pipeline formulation, but improves the frame identification of \citet{hermann2014semantic} with better syntactic features.
\sesame~\citep{swayamdipta2017frame} uses predicted frames from \citet{fitzgerald2015semantic},
and improves argument identification using a softmax-margin segmental RNN.
They observe further improvements from product of experts ensembles \citep{hinton2002training}.

The best published FN 1.5 results are due to~\citet{yang2017joint}. 
Their relational model (\textsc{Rel}) formulates argument identification
as a sequence of local classifications.
They additionally introduce an ensemble method (denoted as \textsc{All}) to
integrate the predictions of a sequential CRF.
They use a linear program to jointly predict frames and arguments at test time.
As shown in Table~\ref{tab:res_15}, our single-model performance
outperforms their \textsc{Rel} model, and is on par with their \textsc{All} model.
For a fair comparison, we build an ensemble (\textsc{Full}, $2\times$) by separately training two models, differing only in random seeds, and averaging their part scores.
Our ensembled model outperforms previous best results by 0.8\% absolute.

\begin{table}[tb]
		\centering
		\begin{tabulary}{\columnwidth}{@{}l   cc@{}}
			\toprule
			
			\textbf{Model}
			& \textbf{All}
			& \textbf{Ambiguous}\\
			
			\midrule
			
			Hartmann
			& 87.6 & -\\
			
			Yang and Mitchell
			& 88.2 & -\\
			
			\midrule[.03em]
			
			Hermann
			& 88.4 & 73.1 \\
			
			\midrule[.03em]
			
			$^\dagger$This work (\textsc{Basic})
			& 89.2 & 76.3\\
			
			$^\dagger$This work (\textsc{NoCTP})
			& 89.2 & 76.4\\
			
			$^\dagger$This work (\textsc{Full})
			& 89.9 & 77.7 \\
			
			$^\dagger$This work (\textsc{Full}, $2\times$)
			& \textbf{90.0} & \textbf{78.0} \\
			
			\bottomrule
		\end{tabulary}
		\caption{Frame identification accuracy on the FN 1.5 test set.
			\textit{Ambiguous} evaluates only on lexical units having more than one possible frames.
			$\dagger$ denotes joint frame and argument identification,
			and bold font indicates best performance.\protect\footnotemark}
		\vspace{-.4cm}
		\label{tab:frame_id}
\end{table}
\footnotetext{Our comparison to \citet{hermann2014semantic} is based on their 
	updated version: \url{http://www.aclweb.org/anthology/P/P14/P14-1136v2.pdf}. 
	Ambiguous frame identification results by \citet{yang2017joint} and \citet{hartmann2017ood}
	are 75.7 and 73.8.
	Their ambiguous lexical unit sets are different 
	from the one extracted from the official frame directory, 
	and thus the results are not comparable to those in Table~\ref{tab:frame_id}.}

Table~\ref{tab:frame_id} compares our frame identification results with previous approaches.
\citet{hermann2014semantic} and \citet{hartmann2017ood} 
use distributed word representations and syntax features.
We follow the \textsc{Full Lexicon} setting~\citep{hermann2014semantic} 
and extract candidate frames from the official directories.
The \textit{Ambiguous} setting compares lexical units with more than one possible frames.
Our approach improves over all previous models under both settings, demonstrating a clear benefit from joint learning.

We observe similar trends on FN 1.7 for both full structure extraction and for frame identification only (Table~\ref{tab:res_17}).
FN 1.7 extends FN 1.5 with more consistent annotations.
Its test set is different from that of FN 1.5, 
so the results are not directly comparable to Table~\ref{tab:res_15}.
We are the first to report frame-semantic parsing results on FN 1.7, and we
encourage future efforts to do so as well.

\begin{table}[tb]
		\centering
		\begin{tabulary}{\columnwidth}{@{}l   rr  r  c r r@{}}
			\toprule
			& 
			\multicolumn{3}{C{0.33\columnwidth} @{}}{\textbf{Full Structure}}
			&& \multicolumn{2}{C{0.22\columnwidth} @{}}{\textbf{Frame Id.}}\\
			\cmidrule{2-4}
			\cmidrule{6-7}
			\textbf{Model}
			& \textbf{Prec.} & \textbf{Rec.} & \textbf{$\bm{F_1}$}
			&& \textbf{All} & \textbf{Amb.}\\
			
			\midrule
			
			\textsc{Basic}
			& 78.0 & 72.1 & 75.0
			&& 88.6 & 76.6 \\
			
			\textsc{NoCTP}
			& 79.8 & 72.4 & 75.9
			&& 88.5& 76.3\\
			
			\textsc{Full}
			& \textbf{80.2} & \textbf{72.9} & \textbf{76.4}
			&& \textbf{89.1} & \textbf{77.5} \\
			
			\bottomrule
		\end{tabulary}
		\caption{FN 1.7 full structure extraction and frame identification test results.
			Bold font indicates best performance.
			FN 1.7 test set is an extension of FN 1.5 test, hence the results here are not comparable to those reported in Table~\ref{tab:res_15}.}
		\label{tab:res_17}
\end{table}

\paragraph{Semantic dependency parsing results.}

\begin{table}[tb]
	\centering
	\begin{tabulary}{\columnwidth}{@{}l  rr@{}}
		\toprule
		
		\textbf{Model}
		& \textbf{id $\bm{F_1}$}
		& \textbf{ood $\bm{F_1}$}\\
		
		\midrule
		
		NeurboParser (\textsc{Basic})
		& 89.4 & 84.5 \\
		
		NeurboParser (\textsc{Freda3}) 
		& 90.4 & 85.3 \\
		
		\midrule[.03em]
		
		NeurboParser (\textsc{Basic}, reimpl.)
		& 90.0 & 84.6 \\
		
		\midrule[.03em]
		
		This work (\textsc{NoCTP})
		& 89.9 & 85.2\\ 
		This work (\textsc{Full})
		& 90.5 & 85.9 \\ 
		This work (\textsc{Full}, $2\times$)
		& \textbf{91.2} & \textbf{86.6} \\ 
		
		\bottomrule
	\end{tabulary}
	\caption{Labeled parsing performance in $F_1$ score for DM semantic dependencies.
		\textit{id} denotes in-domain WSJ test data, and \textit{ood} denotes out-of-domain brown corpus test data.
		Bold font indicates best performance.
	}
	
	\vspace{-.5cm}
	\label{tab:sdp}
\end{table}

Table~\ref{tab:sdp} compares our semantic dependency parsing performance on DM with the baselines.
Our reimplementation of the \textsc{Basic} model slightly improves performance on in-domain test data.
The \textsc{NoCTP} model ties parameters from word embeddings and LSTMs when training on FrameNet and
DM, but does not use cross-task parts or joint prediction.
\textsc{NoCTP} achieves similar in-domain test performance, 
and improves over \textsc{Basic} on out-of-domain data.
By jointly predicting FrameNet structures and semantic dependency graphs, 
the \textsc{Full} model outperforms the baselines by more than 0.6\% absolute $F_1$ scores under both settings.

Previous state-of-the-art results on DM are due to the joint learning model of \citet{peng2017deep}, 
denoted as NeurboParser~(\textsc{Freda3}).
They adopted a multitask learning approach, jointly predicting three different parallel semantic dependency annotations.
Our \textsc{Full} model's in-domain test performance is on par with \textsc{Freda3},
and improves over it by 0.6\% absolute $F_1$ on out-of-domain test data.
Our ensemble of two \textsc{Full} models achieves a new state-of-the-art in
both in-domain and out-of-domain test performance.

\subsection{Analysis}
\begin{table}[tb]
	\small
	\center
	\begin{tabulary}{\columnwidth}{@{}L{0.22\columnwidth}@{\hspace*{0.1cm}}L{0.53\columnwidth}@{\hspace*{0.1cm}}R{0.115\columnwidth}@{\hspace*{0.1cm}}R{0.1\columnwidth} @{}}
		\toprule
		& 
		& \multicolumn{2}{L{0.22\columnwidth} @{}}{\hspace{-.2cm}\textbf{Rel. Err. (\%)}}\\
		\cmidrule{3-4}
		
		\textbf{Operation}
		& \textbf{Description}
		& \textsc{Basic}
		& \textsc{Full}\\
		
		\midrule
		
		\multirow{1}{*}{Frame error}
		& \multirow{1}{\hsize}{Frame misprediction.}
		& \multirow{1}{\hsize}{11.3}& \multirow{1}{\hsize}{11.1}\\
		\hline
		
		\noalign{\smallskip}
		\multirow{2}{*}{Role error}
		& \multirow{2}{\hsize}{Matching span with incorrect role.}
		& \multirow{2}{\hsize}{12.6 (5.2)}& \multirow{2}{\hsize}{13.4 (5.9)}\\
		& & &\\
		\hline
		
		\noalign{\smallskip}
		\multirow{2}{*}{Span error}
		& \multirow{2}{\hsize}{Matching role with incorrect span.}
		& \multirow{2}{\hsize}{11.4}& \multirow{2}{\hsize}{12.3}\\
		& & &\\
		\hline
		
		\noalign{\smallskip}
		\multirow{2}{*}{Arg. error}
		& \multirow{2}{\hsize}{Predicted argument does not overlap with any gold span.}
		& \multirow{2}{\hsize}{18.6}& \multirow{2}{\hsize}{22.4}\\
		& & &\\
		\hline
		
		\noalign{\smallskip}
		\multirow{2}{*}{Missing arg.}
		& \multirow{2}{\hsize}{Gold argument does not overlap with any predicted span.}
		& \multirow{2}{\hsize}{43.5}& \multirow{2}{\hsize}{38.0}\\
		& & &\\
		
		\bottomrule
	\end{tabulary}
	\caption{Percentage of errors made by \textsc{Basic} and \textsc{Full} models on the FN 1.5 development set.
		Parenthesized numbers show the percentage of role errors when frame predictions are correct.
		}
	\label{tab:error_breakdown}	
	\vspace{-.3cm}
\end{table}

\paragraph{Error type breakdown.}
Similarly to \citet{he2017deep},
we categorize prediction errors made by the \textsc{Basic} and \textsc{Full} models 
in Table~\ref{tab:error_breakdown}.
Entirely missing an argument
accounts for most of the errors for both models, 
but we observe fewer errors by \textsc{Full} compared to \textsc{Basic} in this category.
\textsc{Full} tends to predict more arguments in general, including more incorrect arguments. 

Since candidate roles are determined by frames, frame and role errors are highly correlated.
Therefore, we also show the role errors when frames 
are correctly predicted (parenthesized numbers in the second row).
When a predicted argument span matches a gold span,
predicting the semantic role is less challenging.
Role errors account for only around 13\% of all errors,
and half of them are due to mispredictions of frames.

\begin{figure}
	\centering
	\includegraphics[clip,trim=0.cm 0.cm 0.cm 0.cm, width=1.05\columnwidth]{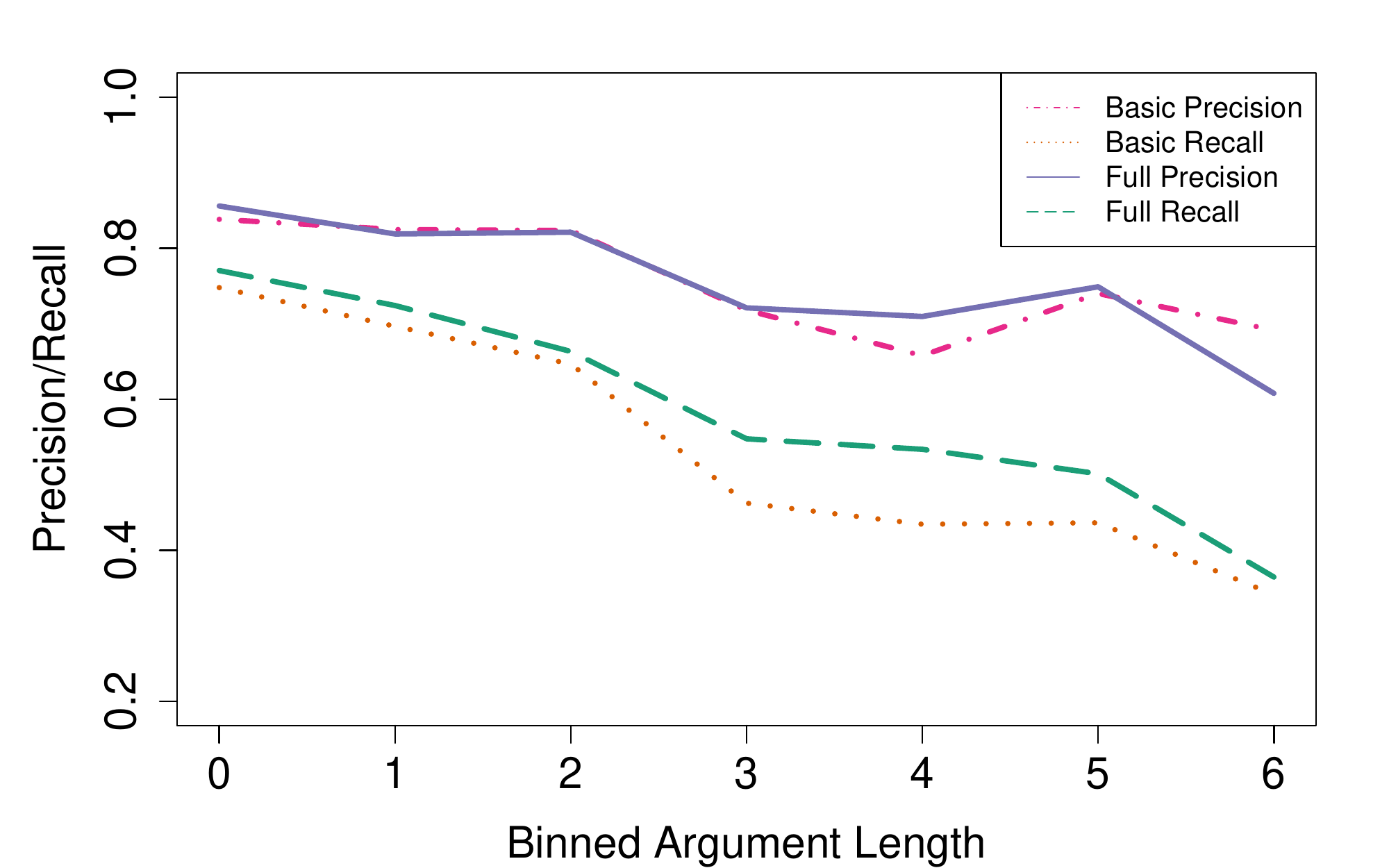}
	\caption{FN 1.5 development precision and recall of \textsc{Basic} and \textsc{Full} by different argument lengths.
		Length $\ell$ is binned to $\floor{\log_{1.6}{\ell}}$, and precision/recall values are smoothed with \texttt{loess}, with a smoothing parameter of 0.1.}
	\label{fig:performance_by_length}
	\vspace{-.3cm}
\end{figure}

\paragraph{Performance by argument length.}
Figure~\ref{fig:performance_by_length} plots dev. precision and recall of both \textsc{Basic} and \textsc{Full}
against binned argument lengths.
We observe two trends: 
(a) \textsc{Full} tends to predict longer arguments (averaging 3.2) compared to \textsc{Basic} (averaging 2.9), 
while keeping similar precision;\footnote{Average gold span length is 3.4 after discarding those longer than 20.}
(b) recall improvement in \textsc{Full} mainly comes from arguments longer than 4. 

\subsection{Implementation Details}
\label{subsec:impl}
Our implementation is based on DyNet \citep{dynet}.\footnote{\url{https://github.com/clab/dynet}}
We use predicted part-of-speech tags and lemmas using NLTK \citep{bird2009natural}.\footnote{\url{http://www.nltk.org/}}

Parameters are optimized with stochastic subgradient descent for up to 30 epochs, 
with $\ell_2$ norms of gradients clipped to 1.
We use 0.33 as initial learning rate, and anneal it at a rate of $0.5$ every 10 epochs.
Early stopping is applied based on FN development ${F}_1$.
We apply logarithm with base 2 to all discrete features, e.g., 
$\log_2(d + 1)$ for distance feature valuing $d$.
To speed up training, we randomly sample a 35\% subset 
from the FN exemplar instances each epoch.

\paragraph{Hyperparameters.}
Each input token is represented as the concatenation a word embedding vector, a learned lemma vector, 
and a learned vector for part-of speech, all updated during training.
We use 100-dimensional \texttt{GloVe} \citep{pennington2014glove} to initialize word embeddings.
We apply word dropout \citep{iyyer2015deep} and randomly replace a word $w$ with a special \texttt{UNK} symbol with probability $\frac{\alpha}{1 + \#(w)}$, with $\#(w)$ being the count of $w$ in the training set.
We follow the default parameters initialization procedure by DyNet, 
and an $\ell_2$ penalty of $10^{-6}$ is applied to all weights. 
See Table~\ref{tb:hyperparameters} for other hyperparameters.

\begin{table}
	\centering
	\begin{tabulary}{\columnwidth}{@{}l r@{}}
		
		\toprule
		
		\textbf{Hyperparameter} & \textbf{Value}\\
		\midrule
		Word embedding dimension & 100 (32) \\
		Lemma embedding dimension & 50 (16) \\
		POS tag embedding dimension & 50 (16) \\
		MLP dimension & 100 (32) \\
		Tensor rank $r$ & 100 (32)\\
		BiLSTM layers & 2 (1)\\
		BiLSTM dimensions & 200 (64)\\
		$\alpha$ for word dropout & 1.0 (1.0)\\
		\bottomrule
	\end{tabulary}
	\caption{Hyperparameters used in the experiments.
			 Parenthesized numbers indicate those used by the pretrained pruners.}
	\label{tb:hyperparameters}
	\vspace{-.5cm}
\end{table}

\paragraph{Modifications to~\citet{peng2017deep}.}
To ensure fair comparisons, we note two implementation modifications
to Peng et al.'s basic model.
We use a more recent version (2.0) of the DyNet toolkit, 
and we use 50-dimensional lemma embeddings instead of 
their 25-dimensional randomly-initialized learned word embeddings.

\section{Conclusion}
\label{sec:conclusion}
We presented a novel multitask approach to learning semantic parsers from
disjoint corpora with structurally divergent formalisms.
We showed how joint learning and prediction can be done with scoring functions
that explicitly relate spans and dependencies, even when they are never
observed together in the data.
We handled the resulting inference challenges with a novel adaptation of
graphical model structure learning to the deep learning setting.
We raised the state-of-the-art on DM and FrameNet parsing by learning from
both, despite their structural differences and non-overlapping data.
\swabha{A line about what we find from the analysis.}
While our selection of factors is specific to spans and dependencies,
our general techniques could be adapted to work with more combinations of
structured prediction tasks.
We have released our implementation at~\url{https://github.com/Noahs-ARK/NeurboParser}.

\section*{Acknowledgments}
We thank Kenton Lee, Luheng He, and Rowan Zellers for their helpful comments,
and the anonymous reviewers
for their valuable feedback.
This work was supported in part by
NSF grant IIS-1562364.
\bibliography{naacl2018}
\bibliographystyle{acl_natbib}

\end{document}